%% file: conference_101719.tex
\documentclass[conference]{IEEEtran}
\IEEEoverridecommandlockouts
\usepackage{cite}
\usepackage{amsmath,amssymb,amsfonts}
\usepackage{algorithmic}
\usepackage{graphicx}
\usepackage{textcomp}
\usepackage{xcolor}
\usepackage{bbm}
\usepackage{verbatim}

\DeclareMathOperator*{\argmax}{argmax}
\def\BibTeX{{\rm B\kern-.05em{\sc i\kern-.025em b}\kern-.08em
    T\kern-.1667em\lower.7ex\hbox{E}\kern-.125emX}}
\begin{document}

\title{A Confidence-Calibrated MOBA Game Winner Predictor
}

\author{\IEEEauthorblockN{Dong-Hee Kim}
\IEEEauthorblockA{\textit{Dept. of Electronics and Computer Engineering} \\
\textit{Hanyang University}\\
Seoul, South Korea \\
dongheekim@hanyang.ac.kr}
\and
\IEEEauthorblockN{Changwoo Lee}
\IEEEauthorblockA{\textit{IUCF} \\
\textit{Hanyang University}\\
Seoul, South Korea \\
cwl30@hanyang.ac.kr}
\and
\IEEEauthorblockN{Ki-Seok Chung}
\IEEEauthorblockA{\textit{Dept. of Electronics and Computer Engineering} \\
\textit{Hanyang University}\\
Seoul, South Korea \\
kchung@hanyang.ac.kr}
}

\IEEEpubid{\begin{minipage}{\textwidth}\ \\[12pt]
978-1-7281-4533-4/20/\$31.00 \copyright 2020 IEEE
\end{minipage}}

\maketitle

\begin{abstract}
In this paper, we propose a confidence-calibration method for predicting the winner of a famous multiplayer online battle arena (MOBA) game, League of Legends.
In MOBA games, the dataset may contain a large amount of input-dependent noise; not all of such noise is observable.
Hence, it is desirable to attempt a confidence-calibrated prediction. Unfortunately, most existing confidence calibration methods are pertaining to image and document classification tasks where consideration on uncertainty is not crucial.
In this paper, we propose a novel calibration method that takes data uncertainty into consideration.
The proposed method achieves an outstanding expected calibration error (ECE) (0.57\%) mainly owing to data uncertainty consideration, compared to a conventional temperature scaling method of which ECE value is 1.11\%.
\end{abstract}

\begin{IEEEkeywords}
Esports, MOBA game, League of Legends, Winning Probability, Confidence-Calibration
\end{IEEEkeywords}
\section{Introduction}

League of Legends (LoL) is arguably one of the most popular multiplayer online battle arena (MOBA) games in the world. It is the game in which the red team and the blue team compete against each other to destroy the opponent's main structure first.
It was reported that 
the 2019 LoL World Championship was the most watched esports event in 2019\cite{starkey_2019}. 
Thus, from both academia and industry, forecasting the outcome of the game in real time has drawn lots of attention.
However, it is a challenging problem to predict the actual winning probability because the outcome of the game may change due to various reasons. 

Many existing efforts to predict the match outcome of the MOBA game in real time have employed machine learning techniques to attain a good prediction accuracy \cite{yang2016real, silva2018continuous, hodge2019win}.
However, we claim that focusing on achieving accuracy may not be adequate for the eSports winner prediction; instead, the predictor should be able to calculate the actual winning probability.
To achieve this goal, \textit{confidence calibration} of neural networks\cite{guo2017calibration} should be taken into consideration. 
Well-calibrated confidence leads to more bountiful and intuitive interpretations of a given data point. For instance, it may not be meaningful to predict the winning team when the predicted winning probability is around 50\%.

\begin{figure}[htbp]
\begin{center}
\centerline{\includegraphics[scale=0.3]{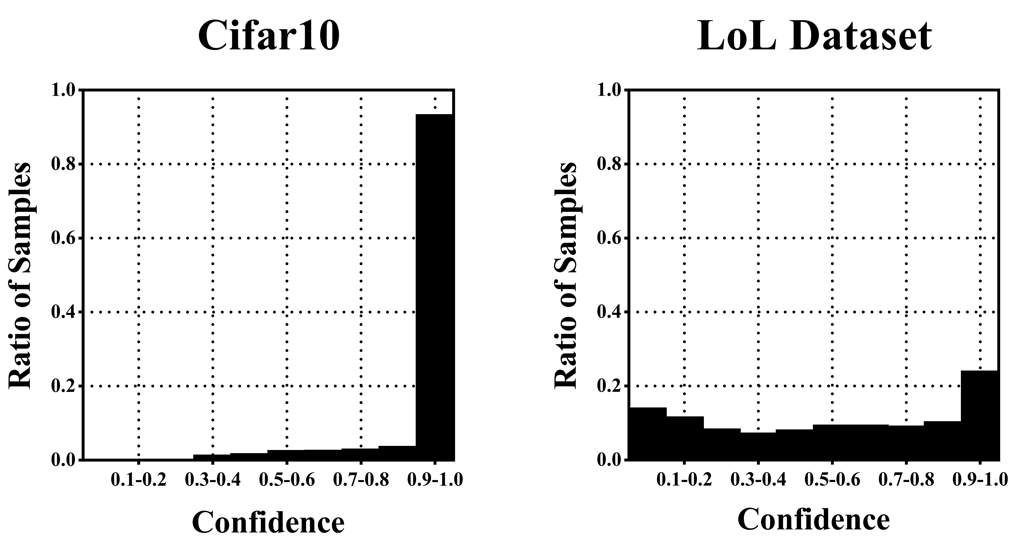}}
\caption{Confidence histograms of Cifar10 and LoL dataset.}
\label{fig:ofSamples}
\end{center}
\end{figure}

How to avoid poorly calibrated models has drawn lots of attention among machine learning researchers. Especially, deep neural networks for image and document classification are mostly overconfident, and there have been several studies to overcome this overconfidence issue\cite{guo2017calibration, seo2019learning}.
From our experiments, we have observed that for over 90\% of samples, the confidence level has turned out to be over 90\% when we trained Cifar10\cite{krizhevsky2009learning} by ResNet-50\cite{he2016deep} as shown in Fig.~\ref{fig:ofSamples}. Therefore, several attempts to train a confidence-calibrated neural network were made to overcome such overconfidence issue.

To train a neural network to predict the outcome of the game, over 93000 matches in-game data per minute have been collected.
Because of the uncertainty of the game, the confidence distribution of the LoL datasets seems evenly distributed contrary to Cifar10 as shown in Fig. \ref{fig:ofSamples}. Owing to this characteristic, not only overconfidence but also underconfidence should become a concern. This concern makes it less effective to adopt conventional calibration methods to the LoL dataset. Hence, how to tame the uncertainty in the LoL data is claimed to be the key for predicting the outcome of LoL.

In this paper, we propose a data-uncertainty taming confidence-calibration method for the winning probability estimation in the LoL environment. The outcome of the proposed method will be the winning probability of a certain team in the LoL game. In Section 2, we will briefly address the evaluation metrics and introduce some existing methods. In Section 3, we propose a data-uncertainty loss function and we will show how it makes the model well-calibrated. We describe the dataset we used and present experimental results in Sections 4. Finally, Section 5 concludes this paper.

\section{Background}
In this section, we look over commonly used calibration metrics and related existing methods. Typically, reliability diagrams are used to show the calibration performance that corresponds to differences between the accuracy and the confidence level. The differences are evaluated in terms of either expected calibration error (ECE), maximum calibration error (MCE), or negative log-likelihood (NLL)\cite{guo2017calibration}. These metrics will have lower values for better calibrated neural networks. The calibration method that we adopt is Platt scaling \cite{platt1999probabilistic}, which is a straightforward, yet practical approach for calibrating neural networks.

First of all, we divide the dataset into $M$ bins based on the confidence level. The interval of the $m^{th}$ bin is expressed as $I_m=\left(\frac{m-1}{M}, \frac{m}{M}\right]$ and we denote the set of samples of which predicted confidence level belongs to $I_m$ as $B_m$. We formally compute the accuracy and the confidence level as:

\begin{equation}
\begin{split}
    \mathrm{acc}(B_m)=\frac{1}{\left\vert B_m \right\vert}\sum_{i: x_i \in B_m} \mathbbm{1}(\hat{y}_i=y_i)\\
    \mathrm{conf}(B_m)=\frac{1}{\left\vert B_m \right\vert}\sum_{i: x_i \in B_m} \hat{p_i} \qquad
\end{split}
\end{equation}
where $\mathbbm{1}$ is an indicator function, $y_i$ and $\hat{y_i}$ denote the true label and the predicted label of the $i^{th}$ sample, respectively, and $\hat{p_i}$ denotes the predicted confidence of the $i^{th}$ sample. The reliability diagram plots the confidence level and the accuracy in one graph. Also, from the accuracy and the confidence level of each bin, ECE and MCE can be calculated where ECE and MCE represent the average error and the maximum error, respectively. ECE and MCE are computed as:
\begin{equation}
\begin{split}
    \mathrm{ECE}=\sum_{m=1}^M\frac{\left\vert B_m \right\vert}{n} \left\vert \mathrm{acc}(B_m)-\mathrm{conf}(B_m) \right\vert\\
    \mathrm{MCE}=\max_{m \in \{1,...,M\}} \left\vert \mathrm{acc}(B_m)-\mathrm{conf}(B_m) \right\vert
\end{split}
\end{equation}
NLL is another way to measure the calibration performance and it is computed as:
\begin{equation}
    \mathrm{NLL}=-\sum_{i=1}^{n}\log p(y_i|x_i,\theta)
\end{equation}

Platt scaling \cite{platt1999probabilistic}, which is a parametric calibration method, shows good calibration results at image and document classification tasks\cite{guo2017calibration}. 
A method called \textit{Matrix scaling} is an extension to the Platt scaling method. In Matrix scaling, logits vector $\mathbf{z}_i$, which denotes raw outputs of a neural network, is generated from the trained model. Then, a linear transformation is applied to the logits to compute the modified confidence level  $\hat{q}_i$ and  the predicted label $\hat{y}^\prime_i$, respectively, as follows:
\begin{equation}
\begin{aligned}
    \hat{q}_i & = \max_k\mathrm{Softmax}(\mathbf{W}\mathbf{z}_i+\mathrm{b})^{(\mathrm{k})}\\
    \hat{y}^\prime_i & = \argmax_k(\mathbf{W}\mathbf{z}_i+\mathrm{b})^{(\mathrm{k})}
\end{aligned}
\end{equation}
where $k$ is the number of classes. 
We optimize matrix $\mathbf{W}$ and bias $\mathrm{b}$ in such a way that the NLL value on the validation set is minimized. \textit {Vector scaling} is similar to Matrix scaling, but it is slightly different in the sense that  $\mathbf{W}$ is restricted to be a diagonal matrix and $\mathrm{b}$ is set to be 0. 

\textit{Temperature scaling} is a simplified version of Platt scaling method in the sense that it relies only on a single positive parameter $T$. Temperature scaling scales a ogits vector $\mathbf{z}_i$ as follows:
\begin{equation}
    \hat{p}_i = \max_k\mathrm{Softmax}(\mathbf{z}_i/T)^{(\mathrm{k})}
\end{equation}
Again, the best $T$ is found in such a way that the NLL value on the validation set is minimized. When $T>1$, the temperature scaling method tries to avoid overconfidence by making the softmax smaller.

\section{Proposed Method}

The aforementioned Platt scaling is dependent only on logits. Therefore, it will generate the same scaling result for two different inputs as long as the two inputs have the same logits.
This limitation is not a problem when the confidence level is quite high as in the case of image classification shown in Fig. \ref{fig:ofSamples}. For the LoL outcome prediction, however, the uncertainty that is inherent in the input data should be taken into consideration when logits are scaled for calibration. Thus, we propose a novel uncertainty-aware calibration method for training a confidence-calibrated model. 

\subsection{Data uncertainty loss function}

\begin{figure*}
\begin{center}
\includegraphics[width=0.85\linewidth]{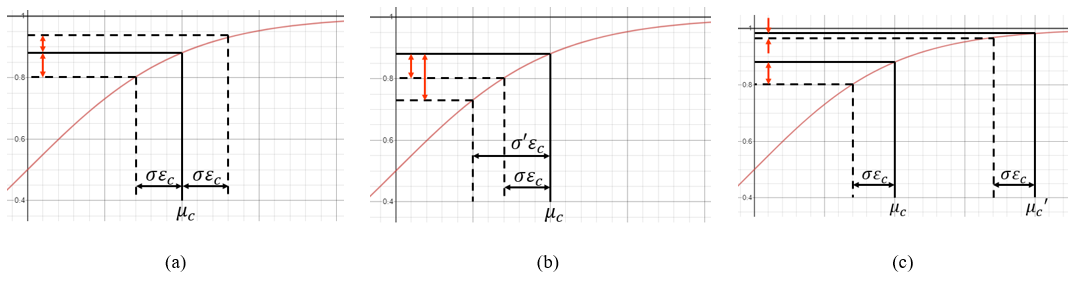}
\end{center}
\caption{Illustration of the sigmoid function to describe how data uncertainty loss function makes the neural network well-calibrated. Red arrows denote the winning probability variation of each situation. (a) Alleviation of overconfidence using uncertainty $\sigma$, (b) The adjusted value for calibration is determined by input uncertainty And (c) High confidence input should not be much swayed by uncertainty.}
\label{fig:sigmoids}
\end{figure*}

In this section, we describe how uncertainty in the input data is measured and how the data uncertainty loss function builds a calibrated model. The data uncertainty is defined as the amount of noise inherent in the input data distribution, and it can be computed as follows:

\begin{equation}
    U_d(y|x) = \mathbb{E}[\mathrm{Var}(y|x)]
\end{equation}

One way to measure the data uncertainty is to utilize a model called \textit{density network} \cite{kendall2017uncertainties, lakshminarayanan2017simple}. 
First, we will discuss about regression tasks. Let $\mu(\mathrm{x})$ and $\sigma(\mathrm{x})$ 
denote the mean and the standard deviation of logits from input $\mathrm{x}$, respectively. Since the standard deviation is a nonnegative number, we take an exponential to the raw output $s(\mathrm{x})$ to predict $\sigma(x)$ so that $\sigma(\mathrm{x})$ is defined as $e^{s(\mathrm{x})}$. With $\mu$ and $\sigma$, we assume that $\mathrm{y}$ is approximated as follows:
\begin{equation}
    \mathrm{y} \sim \mathcal{N}(\mu(\mathrm{x}),\sigma(\mathrm{x})^2)
\end{equation}

In our classification, we get the output of the neural network as logits vector $\boldsymbol{\mu}(\mathrm{x})$ and variance $\sigma(\mathrm{x})^2$. Therefore, we estimate the probability $\boldsymbol{\mathrm{p}}$ as follows:
\begin{equation}
    \boldsymbol{\mathrm{u}} \sim \mathcal{N}(\boldsymbol{\mu}(\mathrm{x}),\mathrm{diag}(\sigma(\mathrm{x})^2))
\label{eq:sample_undiff}
\end{equation}
\begin{equation}
    \boldsymbol{\mathrm{p}} = \mathrm{Softmax}(\boldsymbol{\mathrm{u}})
\end{equation}
Still, re-parametering is necessary to have a differentiable loss function because \eqref{eq:sample_undiff} is not differentiable when the backpropagation is carried out.
To induce the loss function to estimate uncertainty, a Monte Carlo integration is used to derive the probability as in\cite{kendall2017uncertainties}: 
\begin{equation}
    \boldsymbol{\mathrm{u}}^{(k)} \sim \boldsymbol{\mu}(\mathrm{x}) + \mathrm{diag}(\sigma(\mathrm{x}))\epsilon, \epsilon \sim \mathcal{N}(0,I)
\label{eq:rep1}
\end{equation}
\begin{equation}
    \mathbb{E}[\boldsymbol{\mathrm{p}}] \approx \frac{1}{K}\sum^{K}_{k=1}\mathrm{Softmax}(\boldsymbol{\mathrm{u}}^{(k)})
\label{eq:rep2}
\end{equation}
where $\boldsymbol{\mathrm{u}}$ is sampled $K$ times.
Consequently, the loss function includes data uncertainty as:

\begin{equation}
    \mathcal{L}_{\mathrm{clf}} = H(y,\mathbb{E}[\boldsymbol{\mathrm{p}}])
\end{equation}
Here, function $H$ computes the cross-entropy.

\subsection{Calibration effects in data uncertainty loss function}

In this section, we explain why this uncertainty-based loss function makes the model better calibrated. Suppose there are two outputs as in a binary classification task:
\begin{equation}
\begin{split}
    \mathrm{u}^{(1)}=\mu_1+\sigma\epsilon_1\\
    \mathrm{u}^{(2)}=\mu_2+\sigma\epsilon_2
\end{split}
\end{equation}
After the $\mathrm{Softmax}$ layer, the probability of class 1 is derived as:
\begin{equation}
\begin{split}
    \mathrm{p}_1 & = \mathrm{exp}(\mathrm{u}^{(1)})/(\mathrm{exp}(\mathrm{u}^{(1)})+\mathrm{exp}(\mathrm{u}^{(2)})) \\
    & = 1/(1+\mathrm{exp}(\mathrm{u}^{(2)}-\mathrm{u}^{(1)})) \\
    & = 1/(1+\mathrm{exp}(-((\mu_1-\mu_2)+\sigma(\epsilon_1-\epsilon_2)))) \\
    & = \mathrm{Sigmoid}(\mu_c+\sigma\epsilon_c)
\label{eq:sigmoid}
\end{split}
\end{equation}
Here, we note that $\mu_c$ is an input-dependent constant and $\epsilon_c$ will have a different value for every sampling.

Fig.~\ref{fig:sigmoids} shows parts of the sigmoid function around output $x=\mu_c$.
The first thing to note is that the data uncertainty loss function can alleviate the concerns related to the confidence level, notably overconfidence. When $\epsilon_c$ is sampled at two points that have the same absolute value with the opposite signs, the probability $\mathrm{p_1}$ does not change equally. In Fig.\ref{fig:sigmoids}, (a) shows how the probability can change with $\mu_c$ and $\sigma \epsilon_c$. Because of the gradient of the sigmoid function is decreasing gradually at $\mu_c>0$, the probability is changing steeper when $\epsilon_c$ is sampled at the negative side compared to the sample at the positive side with the same absolute value.

In our approach, the calibration effect depends on the input's uncertainty. Fig.\ref{fig:sigmoids}-(b) shows another data of which output has $(\mu_c,\sigma\prime)$ and $\sigma\prime>\sigma$. Even if this dataset has the same $\mu_c$ and the larger uncertainty $\sigma\prime$, the logit value is smaller to result in more uncertain results. Temperature scaling, however, changes $\mu_c$ all at once without considering the characteristics of each input.
Another effect of uncertainty $\sigma$' can be seen with respect to different $\mu_c$ values. With $\mu_c>0$, if the logit increases in proportion to $\mu_c\prime$ as shown in Fig.\ref{fig:sigmoids}-(c), the influence of uncertainty is mitigated. It also helps the neural network better calibrated, because the detrimental influence of the uncertainty is tamed.

\begin{figure*}
\begin{center}
\includegraphics[width=0.75\linewidth]{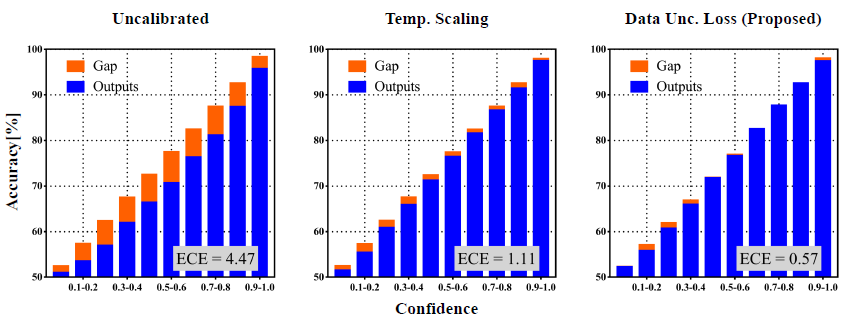}
\end{center}
\caption{Reliability diagrams of the trained network with a League of Legends dataset. The model without calibration shows a large gap between the predicted confidence and the accuracy. On the other hand, a model which is trained with the data uncertainty loss function achieves a remarkably small gap.}
\label{fig:rel_diag}
\end{figure*}



\section{Experimental results}

\begin{table}[htbp]
\caption{Comparison between calibration methods}
\begin{center}
\begin{tabular}{|c|c|c|c|c|}
\hline
 Method         & Accuracy[\%]      & ECE[\%]       & MCE[\%]       & NLL           \\
\hline
No calibration    & 72.95             & 4.47          & 6.76          & 0.515         \\
Temp. Scaling   & 72.95             & 1.11          & 1.87          & \textbf{0.505}\\
Vector Scaling  & 72.71             & 4.63          & 6.66          & 0.536         \\
Matrix Scaling  & 73.05             & 4.43          & 6.90          & 0.540         \\
DU Loss (Proposed)& \textbf{73.81}    & \textbf{0.57} & \textbf{1.26} & 0.515         \\
\hline
\end{tabular}
\label{tabl:ingame}
\end{center}
\end{table}

We implement a [295, 256, 256, 2] shape multi-layer perceptron using Pytorch to predict which team will win. The learning rate chosen at 1e-4 with the Adam optimizer\cite{kingma2014adam} during 20 epochs. The size of all $M$ bins is set to uniformly 10. As the dataset, we collected information on 83875 matches as the train data and that on 10000 matches as the test data. The length of the input vector of the neural network is 295 and it consists of in-game time, champion composition, gold and experience difference and numbers of kills so far. All matches were played between the top 0.01\% players.

The reliability diagram of the trained network with the LoL dataset is shown in Fig.~\ref{fig:rel_diag}. The model without calibration shows the biggest difference between the predicted confidence level and the accuracy across all bins. Especially, for the intervals from $I_5$ to $I_7$, the gap between the confidence level and the accuracy is bigger than 6\%. 
Among all the Platt scaling techniques that we have tried, Temperature scaling turns out to be the best, and our experimental results show that the difference is much smaller with Temperature scaling across all intervals. However, the model that is trained with the proposed calibrated confidence under data uncertainty loss consideration shows almost the same results.

Table \ref{tabl:ingame} summarizes the accuracy for all of the compared calibration methods. For brevity, the calibration with the proposed data uncertainty loss function turns out to be the best method in terms of the accuracy. The model without calibration shows 72.95\% accuracy while Matrix scaling, which shows the best performance among Platt scaling, achieves 73.05\% accuracy. The proposed calibration method achieves the highest accuracy at 73.81\%

We also compared the calibration methods in terms of ECE, MCE and NLL.
The results are summarized in Table \ref{tabl:ingame}. Vector scaling and Matrix scaling have 4.63\% and 4.43\% ECE values, which are worse than other calibration methods. Temperature scaling scores 1.11\% ECE that is the best score among Platt scaling methods. Furthermore, Temperature scaling achieves the best NLL value with 0.505. The proposed method achieves 0.57\% ECE and 1.26\% MCE values, both of them are the best among all compared methods. The model trained with the proposed method achieves a 0.515 NLL result, and it's slightly worse than Temperature scaling, but it is not a significant difference. 

\section{Conclusion}
In this paper, we propose a confidence-calibration method for predicting the winner of a famous multiplayer online battle arena (MOBA) game, League of Legends in real time.
Unlike image and document classification, in MOBA games, datasets contain a large amount of unobservable input-dependent noise. The proposed method takes the uncertainty into consideration to calibrate the confidence level better. 
We compare the calibration capability of the proposed method with commonly used Platt scaling methods in terms of various metrics.
Our experiments verify that the proposed method achieves the best calibration capability in terms of both expected calibration error (ECE) (0.57\%) and maximum calibration error (MCE) (1.26\%) among all compared methods.

\section*{Acknowledgment}
This paper was supported by Korea Institute for Advancement of Technology(KIAT) grant funded by the Korea Government(MOTIE)(N0001883, The Competency Development Program for Industry Specialist)

\bibliographystyle{IEEEtran}
\input{IEEEexample.bbl}

\end{document}